\documentclass[conference]{IEEEtran}

\usepackage[T1]{fontenc}
\usepackage[utf8]{inputenc}
\usepackage{amsmath,amssymb,amsfonts,mathtools}
\usepackage{bm}
\usepackage{upgreek} 
\usepackage{graphicx}
\usepackage{xcolor}
\usepackage{caption}
\usepackage{subcaption}
\usepackage[ruled,linesnumbered]{algorithm2e}
\usepackage{booktabs}
\usepackage{microtype}
\usepackage{cite}
\usepackage{hyperref}
\hypersetup{hidelinks}
\usepackage[capitalize,nameinlink]{cleveref}
\usepackage{mathrsfs}

\newcommand{\vv}[1]{{\boldsymbol{\mathrm{#1}}}}
\newcommand{\mm}[1]{{\boldsymbol{\mathrm{#1}}}}

\newcommand{\bbeta}{{\bm{\upbeta}}}
\newcommand{\bbetah}{\hat{\bbeta}}

\newcommand{\reals}{\mathbb{R}}
\newcommand{\PP}{\mathbb{P}}

\newcommand{\inv}[1]{#1^{-1}}

\newcommand{\yy}{\vv{y}}
\newcommand{\rr}{\vv{r}}
\newcommand{\zz}{\vv{z}}
\newcommand{\ff}{\vv{f}}

\newcommand{\HH}{\mm{H}}
\newcommand{\II}{\mm{I}}

\newcommand{\kp}{\mathrm{kp}}
\newcommand{\meas}{\mathrm{meas}}
\newcommand{\REQ}{\mathrm{REQ}}
\newcommand{\MD}{\mathrm{MD}}
\newcommand{\FA}{\mathrm{FA}}
\newcommand{\NM}{\mathrm{NM}}
\newcommand{\NF}{\mathrm{NF}}
\newcommand{\HMI}{\mathrm{HMI}}
\newcommand{\mon}{\mathrm{mon}}
\newcommand{\PL}{\mathrm{PL}}
\newcommand{\smax}{s_{\max}}
\newcommand{\nkp}{n_\kp}
\newcommand{\proj}{\mathrm{proj}}

\DeclareMathAlphabet{\dutchcal}{U}{dutchcal}{m}{n}
\SetMathAlphabet{\dutchcal}{bold}{U}{dutchcal}{b}{n}
\DeclareMathAlphabet{\dutchbcal}{U}{dutchcal}{b}{n} 
\begin{document}

\title{Protection Levels for Vision-Based Pose Estimation}

\IEEEoverridecommandlockouts
\author{
\IEEEauthorblockN{
Olivia Beyer Bruvik\IEEEauthorrefmark{2}\IEEEauthorrefmark{1},
Romeo Valentin\IEEEauthorrefmark{2}\IEEEauthorrefmark{1},
Marc R. Schlichting\IEEEauthorrefmark{2},
Don Walker\IEEEauthorrefmark{3},
and Mykel J. Kochenderfer\IEEEauthorrefmark{2}%
\thanks{\IEEEauthorrefmark{1} Indicates equal contribution.}%
\thanks{Corresponding author: Olivia Beyer Bruvik (e-mail: oliviabb@stanford.edu).}%
\thanks{\copyright~2026 IEEE.  Personal use of this material is permitted.  Permission from IEEE must be obtained for all other uses, in any current or future media, including reprinting/republishing this material for advertising or promotional purposes, creating new collective works, for resale or redistribution to servers or lists, or reuse of any copyrighted component of this work in other works. Accepted for publication at the 2026 AIAA DATC/IEEE 45th Digital Avionics Systems Conference (DASC).}}
\IEEEauthorblockA{\IEEEauthorrefmark{2}Department of Aeronautics and Astronautics, Stanford University, Stanford, CA 94305 USA\\
\IEEEauthorrefmark{3}A$^3$ by Airbus LLC, Sunnyvale, CA 94086 USA}}

\maketitle

\begin{abstract}
Vision-based navigation complements Global Navigation Satellite Systems, but certification demands integrity guarantees that account for faulty measurements.
Previous work presented a probabilistic computer vision pipeline for runway-based pose estimation with fault detection inspired by Receiver Autonomous Integrity Monitoring.
This work extends that framework by deriving protection levels, which provide probabilistic bounds on pose error that remain valid under undetected faults.
We present an algorithm for computing protection levels for the nonlinear Perspective-$\bm{n}$-Point problem applied to an aviation setting.
The algorithm covers all six degrees of freedom of the aircraft pose (position and orientation) directly.
We analyze the effect of measurement redundancy, pixel-level prediction uncertainty, and runway distance on the resulting protection levels.
To make the results tangible, we demonstrate tradeoffs in the protection levels on an illustrative runway example.
\end{abstract}


\section{Introduction}\label{sec:intro}

Vision-based navigation systems are a promising complement to Global Navigation Satellite Systems (GNSS) for aircraft approach and landing, particularly in GPS-denied or degraded environments.
Such systems perform vision-based pose estimation (VBPE), using learned computer vision models to recover aircraft pose from visual features; however, these models can fail in subtle ways due to, for example, mislabeled corners, misidentified runway edges, or illumination artifacts, producing faulty pose estimates.
Integration into safety-critical avionics therefore requires not only accurate pose estimates but also integrity monitoring for runtime assurance.
Standard integrity monitoring for GNSS-based aviation includes two components: fault detection and the evaluation of a \emph{protection level} (PL).
A PL is a probabilistic upper bound on pose error that holds with a specified probability, even under undetected faults.

PLs are essential for runtime assurance because they provide a real-time, data-driven approach to evaluating when navigation systems can be trusted.
Integrity is assessed by comparing the computed PL to a pre-defined alert limit that sets the maximum pose error tolerable for safe operation.
Whenever the PL is less than the alert limit, the probability of the true pose error exceeding the alert limit is bounded by the specified integrity risk.
Conversely, whenever the PL is greater than the alert limit, the system is marked as unavailable, allowing for secondary instruments to be depended on for navigation.
Tight PLs are essential, since overly conservative bounds preserve integrity at the cost of availability, the fraction of time the system can safely be used \cite{joerger2014,rtca2004do245}.

\begin{figure}[t]
  \centering
  \includegraphics[width=\columnwidth]{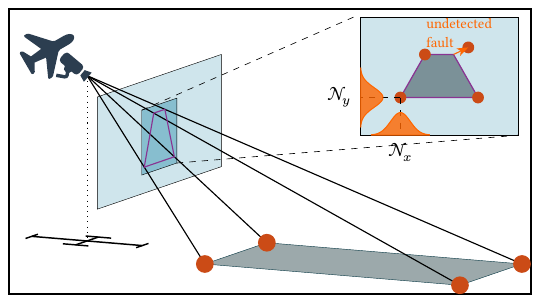}
  \caption{Vision-based pose estimation from runway keypoints; one keypoint prediction contains an undetected fault. We compute probabilistic bounds for the camera pose in this setting.}
  \label{fig:task_overview}
\end{figure}

In GNSS, integrity guarantees are provided by Receiver Autonomous Integrity Monitoring (RAIM) \cite{joerger2014}, which uses measurement redundancy both to detect faults at runtime and to bound the integrity risk through PLs.
Recent work applied RAIM-based fault \emph{detection} to VBPE, providing the first component of runtime assurance for pipelines built around a neural-network keypoint predictor \cite{valentin2025predictive}.
\Cref{fig:task_overview} illustrates the setup: the neural network predicts keypoint pixel locations with per-keypoint uncertainty from an input image, and a Perspective-$n$-Point (PnP) solver recovers the aircraft pose via an optimization problem.

However, that work does not compute PLs, which are the quantitative error bounds that close the integrity argument.
Under Gaussian noise, a nominal pose-error bound follows directly from the propagated pose uncertainty, but aviation requirements demand a bound that also covers undetected faults.
This fault-case bound is currently absent in the VBPE setting.
This paper adapts RAIM's PL construction to supply VBPE PLs for the full six-degree-of-freedom (6-DOF) pose.

We present three contributions.
\begin{enumerate}
  \item We bridge the gap between the VBPE formulation and RAIM theory, allowing us to compute PLs for the full 6-DOF VBPE pose estimation problem.
  \item We analyze the PL equations to characterize key properties in the VBPE setting; specifically, we characterize the effect of distance to runway, keypoint count, and noise magnitude on the size of the PL.
  \item We validate the findings on a representative runway case study and make the results tangible.
\end{enumerate}

The remainder of this paper is organized as follows.
\Cref{sec:related} positions this work against prior literature.
\Cref{sec:prelim} introduces the VBPE problem and reviews the relevant RAIM results in the GNSS setting.
\Cref{sec:methods} derives PLs for VBPE.
\Cref{sec:characteristics} characterizes PL behavior analytically, and \Cref{sec:experiments} validates it on a runway case study.
\Cref{sec:conclusion} concludes and outlines further work.

\section{Related Work}\label{sec:related}

In civil aviation, integrity is the probability that a system is operating correctly or, once faulty, is marked as unavailable. 
The International Civil Aviation Organization (ICAO) and Federal Aviation Administration (FAA) standards make integrity an operational requirement, enforced at runtime using protection levels (PLs).
These PLs upper-bound horizontal and vertical position error at a specified integrity risk, and operations are declared unavailable when a PL exceeds its corresponding alert limit \cite{rtca2004do245}.
This paper focuses on computing such bounds for vision-based aircraft pose estimation.

Classical Perspective-$n$-Point (PnP) methods provide the geometric foundation for this problem.
Solvers such as EPnP \cite{lepetit2009epnp} emphasize computational efficiency and geometric accuracy, while robust methods such as RANSAC \cite{fischler1981ransac} improve resilience to mismatched correspondences.
These methods are essential for pose estimation, but they do not provide an aviation-integrity interpretation of the resulting estimate.
In particular, they do not allocate integrity risk across fault hypotheses, bound the effect of undetected faults, or produce a runtime PL that can be compared with an alert limit.

Uncertainty-aware PnP methods move closer to this objective.
CEPPnP \cite{ferraz2014leveraging} incorporates anisotropic image-observation covariances into a maximum-likelihood PnP solver, and related methods propagate per-keypoint uncertainty to pose-level uncertainty \cite{vakhitov2021uncertainty}.
Set-valued methods provide still stronger guarantees: Yang and Pavone \cite{yang2023conformal} use conformal prediction sets for keypoints and propagate them to pose uncertainty sets, while Luo et al.\ \cite{luo2024certifying} and Santa Cruz and Shoukry \cite{santacruz2023certified} certify vision-based pose-estimation pipelines using verification or reachability analysis.
These approaches provide calibrated or certified uncertainty bounds, but their guarantees are not formulated as aviation-style PLs.
They do not explicitly model monitored fault hypotheses, allocate integrity risk, or tie the resulting bound to an operational alert-limit decision.

The GNSS literature provides the missing integrity framework. Receiver Autonomous Integrity Monitoring (RAIM) uses redundant measurements to detect faults and bound the risk of hazardous misleading information \cite{parkinson1988autonomous,brown1988selfcontained,sturza1988navigation}. Residual-based, parity-based, and solution-separation methods provide different mechanisms for detection and protection-level computation \cite{brown1992baseline,brenner1996integrated,pervan1998multiple,joerger2014,joerger2016fault}. Advanced RAIM extends these ideas to multi-constellation and multi-fault settings by assigning prior fault probabilities, monitoring relevant fault modes, and computing horizontal and vertical PLs for aviation operations \cite{blanch2015baseline,blanch2017protection}. In this literature, PLs are not generic confidence intervals. Instead, they are bounds derived from a fault model, a detection threshold, and an integrity-risk allocation.

Recent work shows that these ideas can be transferred beyond standalone GNSS. Integrity monitoring has been studied for lidar feature association \cite{joerger2016lidar}, multisensor Kalman filtering \cite{meng2021integrity}, sequential monitoring \cite{tanil2018sequential}, camera localization \cite{gupta2021datadriven}, and graph-SLAM with GPS and fisheye cameras \cite{bhamidipati2020integrity}. In aviation, Fu et al.\ \cite{fu2015vision} used visual landmarks as pseudo-satellites to improve GPS integrity monitoring, and Valentin et al.\ \cite{valentin2025predictive} adapted residual-based RAIM for runtime assurance of a vision-based landing system. Zhu et al.\ \cite{zhu2022integrity} review the broader challenges in visual navigation integrity, including environment dependence, association errors, and perception uncertainty.

Together, these results leave a gap. Vision methods increasingly provide robust estimates, covariances, distributions, or calibrated sets, but they generally do not implement the explicit integrity-risk accounting used in aviation. Integrity methods have been extended to perception and multisensor navigation, but existing camera-based examples are largely fusion-based or map-based rather than native learned-keypoint plus nonlinear PnP pipelines. This paper addresses that gap by adapting RAIM-style PL computation to a vision-based pose-estimation pipeline that begins with 2-D keypoint measurements, models keypoint faults, and produces a runtime pose-level bound suitable for comparison with an alert limit.

\section{Preliminaries}\label{sec:prelim}

We briefly review two foundational components that enable computing PLs in the VBPE setting:
(i) a formulation of the probabilistic model for keypoint and pose estimation, and (ii) RAIM in the context of GNSS, which bounds estimation error, even when some measurements are faulty.
In \Cref{sec:methods}, we combine these two approaches.

\subsection{Vision-Based Pose Estimation}\label{sec:vbpe}
We begin by setting up the vision-based pose estimation problem as a \emph{perspective-$n$-point} (PnP) problem.
In this context, the PnP problem aims to estimate the aircraft's pose relative to the runway using an image captured during approach.
This is achieved by estimating the 2-D pixel coordinates of projected keypoints in the image and relating them to known 3-D reference points.
By additionally accounting for the measurement uncertainty of these keypoints, we can construct an optimization problem to robustly compute the camera pose and its associated uncertainty.
The remainder of this section formalizes this procedure.

\subsubsection{PnP Problem}
Let $\bm{\upxi}_k \in \reals^3$ for $k = 1, \ldots, \nkp$ denote a set of known 3-D reference points
\begin{equation}
    \{\bm{\upxi}_k\}_{k=1}^{\nkp} = \{(\mathit{lat}_k, \mathit{lon}_k, \mathit{elev}_k)\}_{k=1}^{\nkp}
\end{equation}
expressed in the world frame.
For example, we can choose the four corners of the runway as reference points, in which case $\nkp = 4$.
However, other features or a larger number of keypoints may be used.

Let $\bbeta^* \in \reals^3 \times \mathrm{SO}(3)$ denote the true aircraft pose, comprising position and orientation.
The estimator operates on an $m$-dimensional state derived from this pose: $m = 6$ for full 6-DOF pose estimation, and $m = 3$ when attitude is provided externally and only position is estimated.
Throughout, a superscript $*$ marks the ground-truth quantities, which are unknown to the estimator.
Given camera intrinsics $\mathcal{C}$ (focal length, principal point, distortion), the projection function then maps each world point to its true pixel location
\begin{equation}\label{eq:projection}
  \vv{y}_k^* = \proj_\mathcal{C}(\bm{\upxi}_k \mid \bbeta^*) \in \reals^2, \quad k = 1, \ldots, \nkp,
\end{equation}
where $\proj_\mathcal{C}$ is the rigid-body transformation from world to camera coordinates followed by perspective projection.
Minimizing the difference between projected and measured keypoint locations provides the foundation of the pose estimation.

\subsubsection{Keypoint Predictions and Noise Model}\label{sec:noise}
In practice, rather than observing pixel projections directly, we receive noisy predictions from a computer vision model such as a neural network.
Crucially, we do not make assumptions about the details of the network architecture.
Instead, given an input image, we assume only that the network outputs a mean pixel location $\hat{{\bm{\upmu}}}_k = (\hat{u}_k, \hat{v}_k)$ and an associated predictive uncertainty $\hat{\boldsymbol{\Sigma}}_k$ for each keypoint $k$.
See \Cref{fig:task_overview} for an overview and \cite{valentin2024probabilistic} for the calibration of these predictive uncertainties.

We interpret these outputs as a nominal, Gaussian measurement model, and departures from it are treated as faults described in \cref{sec:fault}.
To keep the notation simple, we display $\hat{\boldsymbol{\Sigma}}_k$ in diagonal form,
\begin{equation}\label{eq:noise}
  \vv{y}_k \sim \mathcal{N}(\hat{{\bm{\upmu}}}_k, \hat{\boldsymbol{\Sigma}}_k), \quad
  \hat{\boldsymbol{\Sigma}}_k = \begin{pmatrix} \hat{\sigma}_{u_k}^2 & 0 \\ 0 & \hat{\sigma}_{v_k}^2 \end{pmatrix},
\end{equation}
where $\hat{\sigma}_{u_k}$ and $\hat{\sigma}_{v_k}$ are the per-coordinate standard deviations for keypoint $k$.
The method accommodates a general $\hat{\boldsymbol{\Sigma}}_k$ with known correlation.

\subsubsection{Least-Squares Pose Estimation}
Given noisy keypoint predictions, the pose estimate is obtained by minimizing a reprojection residual weighted by the measurement uncertainty.
In the pose estimator, we use the predictive means as the observed keypoint measurements, i.e., $\hat{\vv{y}}_k = \hat{{\bm{\upmu}}}_k$.
Stacking all $2\nkp$ pixel coordinates into a single measurement vector
\begin{equation}
  \hat{\vv{y}} = \begin{pmatrix} \hat{\vv{y}}_1 \\ \vdots \\ \hat{\vv{y}}_{\nkp} \end{pmatrix} \in \reals^{2\nkp},
\end{equation}
the pose estimate $\bbetah$ is obtained by minimizing
\begin{equation}\label{eq:pnp_lsq}
  \bbetah = \arg\min_{\bbeta} \sum_{k=1}^{\nkp} \left\| \hat{\vv{y}}_k - \proj_\mathcal{C}(\bm{\upxi}_k \mid \bbeta) \right\|_{\boldsymbol{\Sigma}_k^{-1}}^2.
\end{equation}
For additional details, see \cite{valentin2025predictive}.
The least-squares problem gives a point estimate $\bbetah$.
For integrity monitoring, we also need the uncertainty of this estimate.

\subsubsection{Computing Pose Uncertainty}\label{sec:state-cov-prelim}
To characterize the uncertainty $\hat{\boldsymbol{\Sigma}}_\bbeta$ of the pose estimate, we linearize the projection function around $\bbetah$ and propagate the per-keypoint uncertainty into the pose space to compute
\begin{equation}\label{eq:state-cov}
  \hat{\boldsymbol{\Sigma}}_\bbeta = (\mm{H}_\mathrm{raw}^\top \boldsymbol{\Sigma}_\meas^{-1} \mm{H}_\mathrm{raw})^{-1},
\end{equation}
where $\boldsymbol{\Sigma}_\meas$ is the block-diagonal covariance $\boldsymbol{\Sigma}_\meas = \mathrm{diag}(\boldsymbol{\Sigma}_1, \ldots, \boldsymbol{\Sigma}_{\nkp})$.
We can now write $\mathcal{N}(\bbetah, \hat{\boldsymbol{\Sigma}}_\bbeta)$ as the predicted probability distribution of the pose.
Here, $\mm{H}_\mathrm{raw} \in \reals^{2\nkp \times m}$ is the raw (unwhitened) observation matrix, obtained by stacking the per-keypoint Jacobians
\begin{equation}
  \mm{H}_{k,\mathrm{raw}} = \left.\frac{\partial \proj_\mathcal{C}({\bm{\upxi}}_k \mid \bbeta)}{\partial \bbeta}\right|_{\bbeta = \bbetah}
\end{equation}
of the projection function at $\bbetah$.
We note that while we assume a Gaussian measurement model and state distribution, other uncertainty propagation techniques are possible that do not assume a Gaussian model, see \cite{valentin2025predictive}.

\subsection{Receiver Autonomous Integrity Monitoring}\label{sec:raim}
The previous section sets up the relationship of the VBPE problem with RAIM.
Next, we review how faulty measurements can be detected.
Given a pose estimate $\bbetah$, we can quantify how likely the observed measurements are under the noise model.
When they are inconsistent with $\bbetah$, at least one keypoint is likely faulty and must be excluded, or the entire measurement set discarded.

RAIM is the standard method for fault detection in GNSS-based aircraft navigation.
In this setting, satellite pseudorange measurements are used to compute position estimates for an aircraft.
The satellite pseudorange measurements are assumed to have zero-mean Gaussian measurement noise, for example due to thermal or multipath effects.
They can also have faults, for example due to satellite clock jumps, spoofing, or other errors not captured by the Gaussian assumption.
Such faults can cause position estimation errors that are unacceptably large for safe navigation.

RAIM provides a framework for fault detection and exclusion by exploiting redundant measurements.
Once a subset of measurements has been selected for which no fault is detected, RAIM can also be used to compute PLs.
In this section, we briefly review the construction of the residual-based (RB) RAIM variant that lays the foundation we adapt to the vision setting presented in \Cref{sec:methods}.

\subsubsection{Measurement Model}
In the GNSS setting, RAIM assumes a linear measurement model derived by linearizing the pseudorange measurement function around a nominal state estimate $\hat{\vv{x}}$:
\begin{equation}\label{eq:RAIM-measurement}
  \zz = \HH \vv{x} + \vv{v} + \vv{f},
\end{equation}
where $\zz = \boldsymbol{\Sigma}^{-1/2} \zz_\mathrm{raw} \in \reals^{2\nkp}$ is the whitened measurement vector, $\HH \in \reals^{2\nkp \times m}$ is the linearized observation matrix, $\vv{x} \in \reals^m$ is the state vector (position and clock bias), $\vv{v} \sim \mathcal{N}(\vv{0}, \II)$ is whitened measurement noise, and $\vv{f} \in \reals^{2\nkp}$ is a fault vector with $\vv{f} = \vv{0}$ under fault-free measurements.

\subsubsection{Fault Detection}
Assuming $\HH$ is full rank and $2\nkp \geq m$, the least-squares estimate of the state $\vv{x}$ is $\hat{\vv{x}} = \HH^\dagger \vv{z}$ with $\HH^\dagger = (\HH^\top \HH)^{-1} \HH^\top$.
Since the true state is not available, RAIM detects faults by comparing the measurement vector $\zz$ to the expected measurement $\HH \hat{\vv{x}}$ at the least-squares estimate $\hat{\vv{x}}$.
This comparison is performed through the residual vector
\begin{equation}\label{eq:residual-vector}
  \vv{r} = \vv{z} - \HH \hat{\vv{x}} = (\II - \HH \HH^\dagger)(\vv{v} + \vv{f}).
\end{equation}

A key insight in RAIM is that the components of $\vv{v} + \vv{f}$ in the column space of $\HH$ are absorbed into the estimate $\hat{\vv{x}}$ and become unobservable to the residual.
Consequently, only components in the orthogonal complement of the column space of $\HH$, known as the left nullspace of $\HH$, remain detectable.
RAIM detects faults from this residual and later bounds the estimation error contributed by the absorbed component.

Recalling \cref{eq:residual-vector}, the residual vector $\vv{r}$ comprises both a noise and a fault component, where $\vv{v} \sim \mathcal{N}(\vv{0}, \II)$.
Assuming a fault-free hypothesis $H_0$ where $\vv{f} = \vv{0}$, the reprojection residual $\vv{r}$ is distributed under a zero-mean Gaussian on the $(2\nkp - m)$-dimensional left nullspace of $\HH$.

Since the squared norm of a standard Gaussian is Chi-squared distributed with degrees of freedom equal to its dimension, and the projection $(\II - \HH \HH^\dagger)$ reduces the ambient $2\nkp$ dimensions to $2\nkp - m$, we have
\begin{equation}\label{eq:q-nofault}
  \|\vv{r}\|_2^2 \sim \chi^2_{2\nkp - m}.
\end{equation}

Considering now a fault hypothesis with $\vv{f} \neq \vv{0}$, $\rr$ now has an additional component induced by the fault vector.
In particular, each element of $\rr$ becomes distributed under a Gaussian with non-zero mean.
Since again we can only observe the component of $\ff$ that lies orthogonal to the column space of $\HH$, we define the non-centrality scale $\lambda^2 = \|(\II - \HH \HH^\dagger) \vv{f}\|_2^2$
and write
\begin{equation}\label{eq:q-fault}
  \|\vv{r}\|_2^2 \sim \chi^2_{2\nkp - m, \lambda^2}.
\end{equation}

\begin{figure}[t]
  \centering
  \includegraphics[width=\columnwidth]{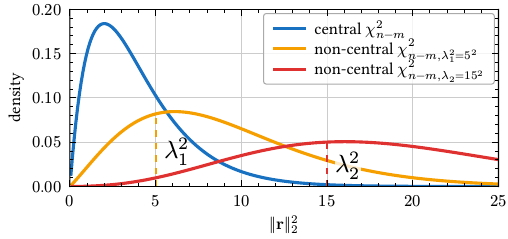}
  \caption{Central $\chi^2_{2\nkp - m}$ density (blue) and two non-central $\chi^2_{2\nkp - m, \lambda^2}$ densities (orange, red) at the same degrees of freedom, with non-centrality scales $\lambda_1 = 5$ and $\lambda_2 = 15$. Dashed lines mark $\lambda_1^2$ and $\lambda_2^2$. As $\lambda$ grows, the fault distribution pulls away from the central one, making it easier to decide whether an observed residual $\|\rr\|_2^2$ was drawn from the nominal or the faulty regime.}
  \label{fig:chisq-distinguishability}
\end{figure}

The problem of fault detection then reduces to the question of whether a particular residual $\|\rr\|_2$ is likely to have been sampled from the central $\chi^2_{2\nkp - m}$ or the non-central $\chi^2_{2\nkp - m, \lambda^2}$ distribution.
\Cref{fig:chisq-distinguishability} illustrates this for two non-centrality scales.

The detector compares the residual norm to a threshold $\tau$, rejecting the measurement if $\|\vv{r}\|_2 > \tau$. In \Cref{sec:pl_derivation}, we will describe how choice of $\tau$ trades false alarms against missed detections and compute it from the integrity-budget allocation.

\subsubsection{Integrity Risk and Protection Levels}
The previous section reviewed fault detection, which has already been adapted to the visual landing application \cite{valentin2025predictive}.
Unfortunately, even when the detection test passes, a fault may still be present, potentially causing an undetected error in the pose estimate that is greater than the specified alert limit $\ell$.

In this case, the system is marked as available and the pose estimate safe to trust despite a pose error that exceeds the alert limit.
This scenario is known as \emph{hazardous misleading information} (HMI) and entails an integrity risk in the system.
We define the integrity risk, or equivalently the probability of HMI, for a state component $j$ as the joint probability
\begin{equation}\label{eq:HMI}
  \PP_\HMI = \PP\bigl(|\hat{x}_j - x^*_j| > \ell \;\wedge\; \|\vv{r}\|_2 < \tau\bigr).
\end{equation}

To bound the probability of HMI, RAIM introduces an integrity risk budget $I_\REQ$ that caps the probability of an undetected estimation error exceeding an alert limit $\ell$.

Notably, faults have different impacts on the estimation error and detectability.
Faults that lie mostly orthogonal to the column space of $\HH$ have higher detectability and less impact on pose error, while faults that lie in the column space of $\HH$ are harder to detect and degrade pose estimates to a greater degree.

Without structural assumptions on $\vv{f}$, the computed PL is infinite: any $\vv{f}$ in the column space of $\HH$ is absorbed into $\hat{\vv{x}}$, shifting the estimate arbitrarily while leaving $\vv{r}$ untouched.
RAIM therefore restricts faults to a discrete set of \emph{fault hypotheses} $H_S$, each modeling a specific failure-mode \cite{joerger2014}.

Since the hypotheses are mutually exclusive and exhaustive, the law of total probability gives the integrity criterion
\begin{multline}\label{eq:integrity}
  \sum_S \PP(H_S)\, \PP\bigl(|\hat{x}_j - x^*_j| > \ell \;\wedge\; \|\vv{r}\|_2 < \tau \mid H_S\bigr) \\
  \leq I_\REQ - \PP_\NM
\end{multline}
for a state component $j$, where $\PP_\NM$ is the unmonitored risk from rare fault combinations that the system does not explicitly check.

A PL is the smallest $\ell$ satisfying \cref{eq:integrity}.
RAIM derives the PL by allocating $I_\REQ$ across the hypotheses and separately bounding each term under the worst case.

\section{Protection Levels for VBPE}\label{sec:methods}
Having reviewed the PnP problem and RAIM in the GNSS setting, we next turn to the main contribution of this paper: computing PLs for the VBPE setting.
The construction proceeds in three subsections: (i) specifying the fault hypotheses (\Cref{sec:fault}), (ii) putting the VBPE measurement model into canonical RAIM form (\Cref{sec:canonical}), and (iii) deriving the per-axis PLs (\Cref{sec:pl_derivation}).

\subsection{Fault Hypotheses}\label{sec:fault}
\begin{figure}[t]
  \centering
  \begin{subfigure}{0.49\columnwidth}
    \centering
    \includegraphics[width=\linewidth]{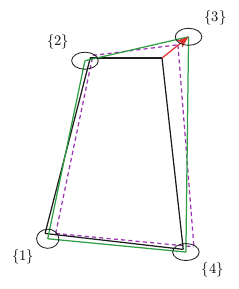}
    \caption{Fault at keypoint 3.}
    \label{fig:runway-fault}
  \end{subfigure}\hfill
  \begin{subfigure}{0.49\columnwidth}
    \centering
    \includegraphics[width=\linewidth]{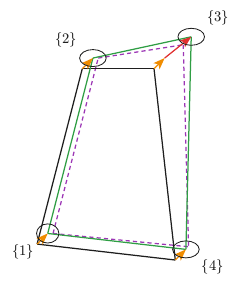}
    \caption{Fault plus noise.}
    \label{fig:runway-fault-noise}
  \end{subfigure}
  \caption{Runway keypoint geometry under correlated noise and a single-keypoint fault. Black: ground truth (GT), {\color[HTML]{2f9e44}green}: predicted keypoints, with $\text{Pred} = \text{GT} + \text{noise} + \text{fault}$; {\color[HTML]{9c36b5}purple} dashed: keypoints reprojected from the LSQ pose estimate. The gap between green and purple is the residual used in detection. {\color[HTML]{f08c00}Orange} arrows: noise (parallel arrows at every keypoint indicate correlated noise); {\color[HTML]{e03131}red} arrow: fault.}
  \label{fig:basic-runway-fault}
\end{figure}

In GNSS RAIM, the fault hypotheses enumerate combinations of pseudorange faults.
In VBPE, these hypotheses enumerate structural errors in the prediction of one or more keypoints.
A structural error is an error that departs from the Gaussian noise model of \cref{eq:noise}.
For example, a fault can occur as a result of a misidentified corner or a misidentified runway edge.
See \Cref{fig:basic-runway-fault} for a visualization.

We index fault hypotheses by the subset $S \subseteq \{1, \ldots, \nkp\}$ of keypoints whose measurements are faulted, with $H_S$ denoting the corresponding hypothesis.
For example, $H_{\{3\}}$ denotes a fault in the third keypoint prediction, $H_{\{3,4\}}$ a simultaneous fault in the third and fourth predictions, and $H_\emptyset \equiv H_0$ the no-fault hypothesis.

A fault hypothesis $H_S$ is \emph{detectable} only when the unfaulted measurements alone are enough to estimate the pose.
Since each keypoint prediction comes with two measurements (horizontal and vertical), we require $2(\nkp - |S|) \geq m$, or equivalently
\begin{equation}\label{eq:redundancy-limit}
  |S| \leq \nkp - m/2.
\end{equation}
In other words, for too many simultaneous and unconstrained faults, RAIM cannot help us detect the presence of a fault.
We therefore monitor only up to $\smax$ simultaneous keypoint faults, i.e., all keypoint sets with at least one and up to $\smax$ keypoint faults.
We define the set of monitored faults
$\dutchcal{S}_\mon = \{ S : 1 \leq |S| \leq \smax \}$.

As a consequence of \cref{eq:redundancy-limit}, for the four-corner runway setup with $\nkp = 4$ and $m = 6$, we can only consider single-keypoint faults ($\smax = 1$).
Monitoring multi-keypoint faults requires additional keypoint predictions such as threshold markers and line angles.

For ease of notation, in the general setting, we assume that $\smax$ is large enough such that hypotheses with $|S| > \smax$ have negligible total prior probability.
This assumption implies an unmonitored risk $\PP_\NM$ small compared to other terms.
We therefore drop it from \cref{eq:integrity} in the rest of this work.

\subsection{Canonical Form}\label{sec:canonical}

To apply the RAIM framework for vision-based pose estimation (VBPE), we manipulate the computer vision measurement function to fit the canonical RAIM form in \cref{eq:RAIM-measurement}.
We construct the canonical form in three steps: linearizing the measurement function, injecting noise and faults, and whitening the result.

\subsubsection{Step 1: Linearize the projection function}
We recall that keypoint projections are computed via the projection function in \cref{eq:projection} and assume that our pose estimate $\bbetah$ is reasonably close to the true pose $\bbeta^*$.
Since the projection function is non-linear, we compute the first-order Taylor expansion for a pose $\bbeta$ near the current pose estimate $\bbetah$.
For each keypoint $k$,
\begin{equation}\label{eq:canonical-derivation-1}
  \proj_\mathcal{C}({\bm{\upxi}}_k \mid \bbeta) \approx \proj_\mathcal{C}({\bm{\upxi}}_k \mid \bbetah) + \mm{H}_{k,\mathrm{raw}}(\bbeta - \bbetah),
\end{equation}
where $\mm{H}_{k,\mathrm{raw}} = \left.\frac{\partial \proj_\mathcal{C}({\bm{\upxi}}_k \mid \bbeta)}{\partial \bbeta}\right|_{\bbeta = \bbetah}$ is the $2 \times 3$ (position) or $2 \times 6$ (pose) Jacobian for keypoint $k$ and the higher-order terms are dropped.

Stacking \cref{eq:canonical-derivation-1} over all $\nkp$ keypoints into the projection
\begin{equation}\label{eq:stacked-proj}
  \proj_\mathcal{C}(\bbeta) =
  \begin{bmatrix}
    \proj_\mathcal{C}(\bm{\upxi}_1 \mid \bbeta) \\
    \vdots \\
    \proj_\mathcal{C}(\bm{\upxi}_{\nkp} \mid \bbeta)
  \end{bmatrix} \in \reals^{2\nkp},
\end{equation}
evaluating at the unknown true pose $\bbeta = \bbeta^*$, and writing the pose offset $\Delta \bbeta := \bbeta^* - \bbetah$ gives
\begin{equation}\label{eq:canonical-derivation-1b}
  \proj_\mathcal{C}(\bbeta^*) - \proj_\mathcal{C}(\bbetah) \approx \mm{H}_\mathrm{raw}\, \Delta \bbeta.
\end{equation}
Recall from \Cref{sec:vbpe} that the keypoint measurement $\hat{\vv{y}}$ contains measurement noise $\vv{v}_\mathrm{raw} \sim \mathcal{N}(\mm{0}, \boldsymbol{\Sigma}_\meas)$, giving the nominal measurement model $\hat{\vv{y}} = \proj_\mathcal{C}(\bbeta^*) + \vv{v}_\mathrm{raw}$.

Defining the measurement deviation vector $\Delta \vv{y} := \hat{\vv{y}} - \proj_\mathcal{C}(\bbetah)$ and substituting the nominal model into \cref{eq:canonical-derivation-1b} yields
\begin{equation}\label{eq:canonical-derivation-2}
  \Delta \vv{y} \approx \mm{H}_\mathrm{raw}\, \Delta \bbeta + \vv{v}_\mathrm{raw}.
\end{equation}

\subsubsection{Step 2: Inject faults}
Under the fault hypothesis $H_S$ (\Cref{sec:fault}), the fault vector takes the form
\begin{equation}\label{eq:fault-selector}
  \vv{f} = \mm{A}_{S,\mathrm{raw}}\, {\bm{\updelta}}_S,
\end{equation}
where the raw fault selector $\mm{A}_{S,\mathrm{raw}} \in \reals^{2\nkp \times 2|S|} = \bigl[\{\vv{e}_{2k-1}, \vv{e}_{2k}\}_{k \in S}\bigr]$ has two columns per faulted keypoint $k \in S$, reflecting that each faulted keypoint corrupts two pixel coordinates simultaneously.
Adding $\mm{A}_{S,\mathrm{raw}}\, {\bm{\updelta}}_S$ to \cref{eq:canonical-derivation-2} yields
\begin{equation}\label{eq:canonical-derivation-3}
  \Delta \vv{y} \approx \mm{H}_\mathrm{raw}\, \Delta \bbeta + \vv{v}_\mathrm{raw} + \mm{A}_{S,\mathrm{raw}}\, {\bm{\updelta}}_S.
\end{equation}

\subsubsection{Step 3: Whiten the noise}
Finally, we bring the formulation into its canonical form by whitening the noise in $\vv{v}$.
We factorize $\boldsymbol{\Sigma}_\meas = \mm{L}\mm{L}^\top$, so that $\inv{\mm{L}}\, \vv{v}_\mathrm{raw} \sim \mathcal{N}(\vv{0}, \II)$.
Multiplying \cref{eq:canonical-derivation-3} by $\inv{\mm{L}}$ on the left gives
\begin{equation}
  \Delta \vv{z} = \HH\, \Delta \bbeta + \vv{v} + \mm{A}_S\, {\bm{\updelta}}_S,
\end{equation}
where $\Delta \zz = \inv{\mm{L}} \Delta \yy$, $\HH = \inv{\mm{L}} \mm{H}_\mathrm{raw}$, and $\mm{A}_S = \inv{\mm{L}} \mm{A}_{S,\mathrm{raw}}$.
Dropping $\Delta$ for ease of notation yields the canonical form
\begin{equation}\label{eq:canonical-form}
  \boxed{\;\vv{z} = \HH\, \bbeta + \vv{v} + \mm{A}_S\, {\bm{\updelta}}_S\;}
\end{equation}
equivalent to \cref{eq:RAIM-measurement}, where
\begin{itemize}
  \item $\vv{z} \in \reals^{2\nkp}$ is the whitened reprojection error, distinct from the residual $\vv{r} = (\II - \HH \HH^\dagger) \vv{z}$ used by the detector;
  \item $\HH \in \reals^{2\nkp \times m}$ is the whitened linearized observation matrix of the projection function $\proj_\mathcal{C}$ around $\hat{\bbeta}$ with $m = 3$ or $6$;
  \item $\bbeta \in \reals^m$ is the pose perturbation;
  \item $\vv{v} = \inv{\mm{L}} \vv{v}_\mathrm{raw} \sim \mathcal{N}(\vv{0}, \II)$ is the whitened noise; and
  \item $\ff = \mm{A}_S\, {\bm{\updelta}}_S$ is the whitened fault under $H_S$, with ${\bm{\updelta}}_S \in \reals^{2|S|}$ and ${\bm{\updelta}}_S = \vv{0}$ under $H_0$.
\end{itemize}
The PL derived below bounds the local axial estimation error in this linearized model.
In the original pose coordinates, this corresponds to a bound on $|\hat{\bbeta}_j - \bbeta^*_j|$ up to the linearization approximation.

\subsection{Protection Level Computation}\label{sec:pl_derivation}

With the fault model and canonical form established, we compute PLs for the VBPE setting.
We assume the computer vision model has produced an initial set of measurements $\{\hat{\vv{y}}_k\}_{k=1}^{\nkp}$ that pass the residual-based detection test described in \Cref{sec:raim}.
See \cite{valentin2025predictive} for further details.\footnote{If the measurements do not pass the residual test, fault detection and exclusion (FDE) may be applied before computing the PL. We leave FDE and its effect on PLs in this setting for future work.}

For each axis $j$ of the pose estimate $\bbetah$, we seek a conservative $\PL_j$ that bounds the probability of hazardous misleading information.
Here, $j$ can represent position or rotation axes.
Following \cref{eq:integrity}, we require
\begin{multline}\label{eq:target-guarantee}
  \sum_{S \in \dutchcal{S}_\mon} \PP(H_S)\, \PP\bigl(|\hat{\bbeta}_j - \bbeta^*_j| > \PL_j \;\wedge\; \|\vv{r}\|_2 < \tau \mid H_S\bigr) \\
  + \PP(H_0)\, \PP\bigl(|\hat{\bbeta}_j - \bbeta^*_j| > \PL_j \mid H_0\bigr) \leq I_\REQ,
\end{multline}
where now the unmonitored risk is dropped.

The process to compute PLs is then as follows.
Each PL has two components:
(i) a bound on the pose error simply arising from the presence of nominal noise ($H_0$), and
(ii) a bound arising from the possibility of an undetected fault ($H_S$).

The nominal-noise component follows directly from the propagated pose covariance.
The fault component is less readily available because we do not know either the direction or the magnitude of a potentially undetected fault.
We therefore account for the worst-case undetected fault as follows.
For each fault hypothesis, we first find the worst-case fault direction that is simultaneously the hardest to detect and the most damaging to the pose estimate.
Then, we find the worst-case fault magnitude along that direction: the largest value for which the residual test could still miss the fault with non-negligible probability.
We then satisfy all hypotheses at once by considering the largest PL among all hypotheses;
this yields the fault-induced PL.
Finally, we conservatively add the nominal component to the fault-induced PL to yield the final PL.
The next sections elaborate on each step.

\subsubsection{Risk Allocation}
We start by detailing how the integrity risk is split up across all fault hypotheses (including the no-fault hypothesis).
Since \cref{eq:target-guarantee} bounds the total probability of HMI under both the fault hypotheses and the no-fault hypothesis, we can simplify the computation by splitting the integrity risk into two parts: $I_\REQ = I_{H_0} + I_{H_\dutchcal{S}}$.
In the absence of calibrated keypoint-fault statistics, we adopt the baseline split
$I_{H_0} = I_{H_\dutchcal{S}} = I_\REQ / 2$.

Equation~\eqref{eq:target-guarantee} is then satisfied if the total probability of HMI in the nominal case is less than $I_{H_0}$, and similarly the total probability of HMI due to faults is less than $I_{H_\dutchcal{S}}$.

Using a similar argument, we further split $I_{H_\dutchcal{S}}$ into hypothesis-level risk allocations as
\begin{equation}\label{eq:fault-risk-allocation}
  I_{H_S} = \frac{\PP(H_S)}{\sum_{S' \in \dutchcal{S}} \PP(H_{S'})}\, I_{H_\dutchcal{S}}
\end{equation}
with $\dutchcal{S} = \dutchcal{S}_\mon$.
Then $I_{H_\dutchcal{S}} = \sum_{S \in \dutchcal{S}} I_{H_S}$ and $I_{H_S} \propto \PP(H_S)$.
Next, we describe how to define $\PP(H_S)$ in the VBPE setting.

Unlike in GNSS, where satellite fault probabilities are supported by historical monitoring data, there is no analogous calibrated fault model for neural-network keypoint predictions.
We assume that keypoint faults occur independently with a common per-keypoint failure probability of $\PP_\kp$.
For an exact fault subset $S \subseteq \{1, \ldots, \nkp\}$, this gives
\begin{equation}\label{eq:product-priors}
  \PP(H_S) = (\PP_\kp)^{|S|} (1 - \PP_\kp)^{\nkp - |S|}
\end{equation}
with a null-hypothesis probability $\PP(H_0) = (1 - \PP_\kp)^{\nkp}$.
We leave empirical calibration of prediction fault probabilities as future work.
Now that we have discussed how to distribute the integrity risk across hypotheses, we proceed with analyzing individual hypotheses.

\subsubsection{Worst-Case Fault Direction}
For hypothesis $H_S$, we now compute the worst-case fault direction.
As introduced in \Cref{sec:canonical}, we write the whitened fault as $\vv{f} = \mm{A}_S\, {\bm{\updelta}}_S$.
The fault component in the column space of $\HH$, calculated as $\HH \HH^\dagger \vv{f}$, is absorbed into the pose estimate, and the remaining component $(\II - \HH \HH^\dagger) \vv{f}$ is detectable in the residual.

For a fixed monitored hypothesis $H_S$, the fault direction $\vv{f}$ is unknown.
To avoid making assumptions about fault distributions, we use the direction that maximally shifts the pose along the axis of interest while remaining hardest to detect.
We capture this trade-off by the \emph{failure-mode slope} \cite[Eq.~30]{joerger2014}
\begin{equation}\label{eq:slope-defn}
  g_{S,j} = \frac{|\vv{e}_j^\top \HH^\dagger \vv{f}|}{\|(\II - \HH \HH^\dagger) \vv{f}\|_2}.
\end{equation}
The numerator is the fault-induced estimate error along axis $j$, extracted by the standard basis vector $\vv{e}_j$.
The denominator is the detectable residual component, equivalently the non-centrality parameter $\lambda = \|(\II - \HH \HH^\dagger) \vv{f}\|_2$ of the residual chi-square under $H_S$.
A large slope means that a fault can substantially move the pose estimate while leaving only a small residual.

Under hypothesis $H_S$, faults are restricted to the range of $\mm{A}_S$.
When $2|S| \leq 2\nkp - m$ and the corresponding fault directions are observable in the residual space, the matrix $\mm{A}_S^\top (\II - \HH \HH^\dagger) \mm{A}_S$ is invertible and the worst-case residual-based RAIM direction has the closed form \cite[Eq.~33]{joerger2014}
\begin{equation}\label{eq:fi}
  \bar{\vv{f}}_S = \mm{A}_S \bigl(\mm{A}_S^\top (\II - \HH \HH^\dagger) \mm{A}_S\bigr)^{-1} \mm{A}_S^\top \HH^{\dagger\top} \vv{e}_j.
\end{equation}

The corresponding worst-case failure-mode slope is
\begin{equation}\label{eq:gfi}
  \bar{g}_{S,j}^2 = \vv{e}_j^\top \HH^\dagger \mm{A}_S \bigl(\mm{A}_S^\top (\II - \HH \HH^\dagger) \mm{A}_S\bigr)^{-1} \mm{A}_S^\top \HH^{\dagger\top} \vv{e}_j.
\end{equation}

\subsubsection{Maximum Undetected Fault Magnitude}
Having computed the worst-case fault direction and the corresponding failure-mode slope, we now turn to the magnitude of the fault.
Concretely, we seek the largest magnitude fault in the worst-case fault direction that goes undetected by the residual test.

Since the measurements passed the fault detection test, we know that the residual $\vv{r} = (\II - \HH \HH^\dagger)(\vv{v} + \vv{f})$ satisfies $\|\vv{r}\|_2 \leq \tau$ for the threshold $\tau$ defined in \Cref{sec:raim}.

The test compares $\|\vv{r}\|_2$ to $\tau$, with $\|\vv{r}\|_2^2 \sim \chi^2_{2\nkp - m}$ under $H_0$ and $\chi^2_{2\nkp - m, \lambda^2}$ under $H_S$, where $\lambda = \|(\II - \HH \HH^\dagger) \vv{f}\|_2$ depends on the fault realization.
The threshold $\tau$ trades off the false-alarm and missed-detection rates: too small a $\tau$ rejects too many nominal samples; too large a $\tau$ admits too many faulty samples.
The detection threshold is set from the continuity allocation $C_\REQ$, which we assume is equal to the probability of false alarm $\PP_\FA$:
\begin{equation}\label{eq:choose-tau}
  \text{find } \tau \text{ s.t. } \PP(\|\rr\|_2 > \tau \mid H_0)\, \PP(H_0) \leq C_\REQ.
\end{equation}

Recalling \Cref{fig:chisq-distinguishability}, under $H_0$, the probability that an observed residual lands to the right of the threshold equals $\PP_\FA$.
Symmetrically, under a fault hypothesis $H_S$, the probability that a residual falls to the left of the threshold equals the probability of missed detection under $H_S$, denoted $\PP_{\MD,S}$.

For a given hypothesis $H_S$, we compute the missed detection probability $\PP_{\MD,S}$ directly from the allocated integrity risk $I_{H_S}$.
Dividing by the probability of $H_S$ and substituting \cref{eq:fault-risk-allocation} into $I_{H_S}$ yields the hypothesis-independent missed detection probability
\begin{equation}\label{eq:pmd}
  \PP_{\MD,S} = \frac{I_{H_S}}{\PP(H_S)} = \frac{I_{H_\dutchcal{S}}}{\sum_{S' \in \dutchcal{S}_\mon} \PP(H_{S'})}.
\end{equation}
Under the choice \cref{eq:fault-risk-allocation}, the right hand side is independent of the concrete hypothesis $S$.
Thus all $\PP_{\MD,S}$ are equal, and we write $\PP_\MD = \PP_{\MD,S}$.

The non-centrality scale $\bar{\lambda}$ is then defined by
\begin{equation}\label{eq:lambda-bar}
  \PP\bigl(\|\vv{r}\|_2 \leq \tau \mid \bar{\lambda}^2\bigr) = \PP_\MD
\end{equation}
with $\|\vv{r}\|_2^2 \sim \chi^2_{2\nkp - m, \bar{\lambda}^2}$, computed by numerical inversion of the non-central $\chi^2$ CDF.
Thus $\bar{\lambda}$ is the detection-boundary non-centrality.
A fault with residual scale $\bar{\lambda}$ is missed with probability $\PP_\MD$.
It depends on the detector threshold, the degrees of freedom, and the fault-budget allocation, but not on the particular monitored hypothesis.

\begin{algorithm}[t]
\DontPrintSemicolon
\caption{PL computation for VBPE.}
\label{alg:pl}
\KwIn{Whitened observation matrix $\HH \in \reals^{2\nkp \times m}$, whitening factor $\inv{\mm{L}}$ from $\boldsymbol{\Sigma}_\meas = \mm{L} \mm{L}^\top$, continuity allocation $C_\REQ$, integrity budget $I_\REQ$, monitored set $\dutchcal{S}_\mon$, maximum monitored fault size $s_{\max}$, priors $\{\PP(H_S)\}_{S \in \{\emptyset\} \cup \dutchcal{S}_\mon}$.}
\KwOut{$\PL_j$ with $j = 1, \ldots, m$.}
Detection threshold: $\tau \gets \sqrt{F_{\chi^2_{2\nkp - m}}^{-1}(1 - C_\REQ / \PP(H_0))}$\;
Unmonitored prior mass: $\PP_\NM \gets \sum_{|S| > \smax} \PP(H_S)$\;
Equal split: $I_{H_0} \gets I_{H_\dutchcal{S}} \gets (I_\REQ - \PP_\NM) / 2$\;
No-fault multiplier: $k_{H_0} \gets \Phi^{-1}\!\bigl(1 - I_{H_0} / (2\, \PP(H_0))\bigr)$\;
Agg.\ missed-detection: $\PP_\MD \gets I_{H_\dutchcal{S}} / \sum_{S \in \dutchcal{S}_\mon} \PP(H_S)$\;
Non-centrality: find $\bar{\lambda}$ s.t.\ $\PP(\chi^2_{2\nkp - m, \bar{\lambda}^2} < \tau^2) = \PP_\MD$\;
State covariance: $\hat{\boldsymbol{\Sigma}}_\bbeta \gets (\HH^\top \HH)^{-1}$\;
\For{$j = 1, \ldots, m$ \textnormal{(each state of interest)}}{
  $\sigma_{\beta_j} \gets \sqrt{\vv{e}_j^\top \hat{\boldsymbol{\Sigma}}_\bbeta \vv{e}_j}$\;
  \For{$S \in \dutchcal{S}_\mon$ \textnormal{(each fault hypothesis)}}{
    $\mm{A}_{S,\mathrm{raw}} \gets [\{\vv{e}_{2k-1}, \vv{e}_{2k}\}_{k \in S}]$\;
    $\mm{A}_S \gets \inv{\mm{L}} \mm{A}_{S,\mathrm{raw}}$\;
    $\bar{g}_{S,j}^2 \gets \vv{e}_j^\top \HH^\dagger \mm{A}_S (\mm{A}_S^\top \mm{P} \mm{A}_S)^{-1} \mm{A}_S^\top \HH^{\dagger\top} \vv{e}_j$ where $\mm{P} = \II - \HH \HH^\dagger$\;
  }
  $\PL_j \gets (\max_{S \in \dutchcal{S}_\mon} \bar{g}_{S,j}) \cdot \bar{\lambda} + k_{H_0} \cdot \sigma_{\beta_j}$\;
}
\Return{$(\PL_1, \ldots, \PL_m)$}\;
\end{algorithm}

\subsubsection{Final Protection Level}
We now combine the fault and nominal components in \cref{eq:target-guarantee} into a PL.
The nominal contribution follows from the propagated pose covariance.
The fault contribution combines the missed-detection non-centrality $\bar{\lambda}$ with the worst-case slope $\bar{g}_{S,j}$ that maps residual bias into axial pose error.
Following residual-based RAIM, we take the worst case over the monitored fault hypotheses.

Under $H_0$, no fault is present, and the axial pose error is modeled as a zero-mean Gaussian with variance $\sigma_{\beta_j}^2 = \vv{e}_j^\top \hat{\boldsymbol{\Sigma}}_\bbeta \vv{e}_j$, where the pose covariance $\hat{\boldsymbol{\Sigma}}_\bbeta$ is computed in \cref{eq:state-cov}.
The probability that this Gaussian noise alone pushes the estimate past a half-width $\ell$ is bounded by the standard tail
\begin{equation}
  \PP(|\hat{\bbeta}_j - \bbeta_j^*| > \ell \mid H_0)\, \PP(H_0) \leq I_{H_0}.
\end{equation}
Inverting at the $H_0$ budget share gives the no-fault contribution
\begin{equation}\label{eq:pl-nf}
  \PL_{0,j} = k_{H_0}\, \sigma_{\beta_j}, \quad k_{H_0} = \Phi^{-1}\!\left(1 - \frac{I_{H_0}}{2\, \PP(H_0)}\right),
\end{equation}
i.e., the standard Gaussian quantile of $\mathcal{N}(0, \sigma_{\beta_j}^2)$ at the integrity-tail level allocated to $H_0$.

For a monitored fault hypothesis $H_S \in \dutchcal{S}_\mon$, the worst undetected fault contributes at most $\bar{g}_{S,j}\, \bar{\lambda}$ of error along axis $j$, the hypothesis-specific slope times the hypothesis-independent non-centrality.
We conservatively add the nominal Gaussian margin, giving
\begin{equation}\label{eq:pl-i}
  \PL_{S,j} = \bar{g}_{S,j} \cdot \bar{\lambda} + k_{H_0} \cdot \sigma_{\beta_j}.
\end{equation}
We recover the $H_0$ contribution \eqref{eq:pl-nf} as $\PL_{\emptyset,j} = k_{H_0} \sigma_{\beta_j}$ with $\bar{g}_{\emptyset,j} = 0$.
The final PL is the maximum over all monitored hypotheses:
\begin{equation}\label{eq:PL-final}
  \PL_j = \max_{S \in \dutchcal{S}_\mon} \PL_{S, j} = \Bigl(\max_{S \in \dutchcal{S}_\mon} \bar{g}_{S,j}\Bigr) \cdot \bar{\lambda} + k_{H_0}\, \sigma_{\beta_j}.
\end{equation}
The resulting $\PL_j$ is a conservative per-axis bound satisfying \cref{eq:target-guarantee}.

\Cref{alg:pl} summarizes the full procedure.
All quantities are computed from the whitened $2\nkp \times m$ observation matrix $\HH$, the noise covariance, and the risk allocations.

\section{Protection Level Characteristics}\label{sec:characteristics}
Having established the PL equations, we aim to understand fundamental characteristics of the computed PLs.
In particular, we analyze the impact of increased observation noise, additional keypoint predictions, and providing known orientation ($m=3$ instead of $m=6$).

To make the analysis tractable, we assume single-keypoint faults ($\smax = 1$), homoscedastic noise $\boldsymbol{\Sigma}_\meas = \sigma_\mathrm{scale}^2 \II$, and the geometric regularity that as $\nkp$ grows, keypoints continue to span the runway rather than clustering throughout this section.

\subsection{Noise magnitude}\label{sec:noise-magnitude}
To start, we recall that the PL formula in \cref{eq:PL-final} separates into a nominal-noise term $k_{H_0}\, \sigma_{\beta_j}$ and a fault term $\bar{g}_{S,j}\, \bar{\lambda}$.
In the following, we denote these components $\PL_\NF$ and $\PL_S$.
It is then straightforward to see that $\PL_\NF$ is proportional to the measurement noise factor $\sigma_\mathrm{scale}$.
Consider that $k_{H_0}$ is only computed from whitened inputs (and therefore independent of $\sigma_\mathrm{scale}$), and recall from \cref{eq:state-cov} that $\boldsymbol{\Sigma}_\bbeta = (\mm{H}_\mathrm{raw}^\top \boldsymbol{\Sigma}_\meas^{-1} \mm{H}_\mathrm{raw})^{-1}$.
From the latter, we can easily see that $\sigma_{\beta_j}$ is directly proportional to $\sigma_\mathrm{scale}$.

For $\PL_S$, the argument is more involved but yields the same result.
We proceed by making the PL formulation fully independent of the noise by right-normalizing $\HH$ with the pose uncertainty, in addition to the left-normalization by the measurement noise in \Cref{sec:raim}.
Specifically, let $\mm{D} = \mathrm{diag}(\hat{\boldsymbol{\Sigma}}_\bbeta)$.
Then, inserting $\mm{D}^{1/2} \mm{D}^{-1/2}$ (an algebraic identity) between $\HH$ and $\bbeta$ in the canonical form \cref{eq:canonical-form} rewrites the model as $\vv{z} = (\HH \mm{D}^{1/2})(\mm{D}^{-1/2} \bbeta) + \vv{v} + \mm{A}_S {\bm{\updelta}}_S$.
Here, $\mm{D}$ rescales each column of $\HH$ by the per-axis nominal standard deviation $\sigma_{\beta_j}$ of the pose estimate.
Under homoscedastic noise, $\HH \propto \sigma_\mathrm{scale}^{-1}$ and $\mm{D}^{1/2} \propto \sigma_\mathrm{scale}$, so the doubly-normalized $\HH \mm{D}^{1/2}$ is $\sigma_\mathrm{scale}$-free.
Thus, using $\HH \mm{D}^{1/2}$ instead of $\HH$ to compute PLs yields unit-free and $\sigma_\mathrm{scale}$-invariant PLs, which relate to $\mm{D}^{-1/2} \bbeta$ instead of $\bbeta$.
Multiplying these unit-free PLs by the $j$-th component of $\mm{D}^{1/2}$, which is exactly $\sigma_{\beta_j}$, recovers the fault-induced PL. Since $\sigma_{\beta_j} \propto \sigma_\mathrm{scale}$, we find that the fault-induced PLs are also proportional to $\sigma_\mathrm{scale}$.

In summary, both terms of \cref{eq:PL-final} scale linearly with the observation noise $\sigma_\mathrm{scale}$, and we write
\begin{equation}
  \PL \propto \sigma_\mathrm{scale}.
\end{equation}

\subsection{Number of keypoints}\label{sec:n-keypoints}
Next, we consider the effect of including additional keypoint measurements.
The two PL terms decay at different rates as $\nkp$ grows.
For $\PL_\NF$, each keypoint contributes two independent rows to $\HH$, so under the regularity assumption, every entry of $\HH^\top \HH$ grows linearly with $\nkp$.
Its inverse, the pose covariance $\hat{\boldsymbol{\Sigma}}_\bbeta = (\HH^\top \HH)^{-1}$, therefore has every entry approximately scaling with $1/\nkp$.
Thus, $\sigma_{\beta_j}^2 \propto \nkp^{-1}$ or, equivalently, $\sigma_{\beta_j} \propto \nkp^{-1/2}$.

The second component in $\PL_\NF$ is the multiplier $k_{H_0}$.
As established in \cref{eq:pl-nf}, $k_{H_0} = \Phi^{-1}(1 - I_{H_0} / (2\, \PP(H_0)))$.
Here, $k_{H_0}$ depends on $\nkp$ only through the no-fault prior $\PP(H_0) = (1 - \PP_\kp)^{\nkp}$ introduced in \cref{eq:product-priors}.
If we assume $\PP_\kp$ small, $\PP(H_0)$ is approximately $1$ for small to moderate $\nkp$.
Hence,
\begin{equation}
  \PL_\NF = k_{H_0}\, \sigma_{\beta_j} \propto \nkp^{-1/2}.
\end{equation}

Next, we consider the fault-induced $\PL_S$.
The fault term is the product $\bar{g}_{S,j}\, \bar{\lambda}$ of the slope and the non-centrality, which scale separately.

For the slope, consider \cref{eq:gfi} but for the double-normalized setting.
Since $\bar{g}_{S,j}^2$ scales with the inverse of $\HH \HH^\dagger$, we find that the double-normalized $\bar{g}_{S,j}^2$ is proportional to $\nkp^{-1}$, and thus $\bar{g}_{S,j} \propto \nkp^{-1/2}$.
However, correcting for the effect of $\sigma_{\beta_j} \propto \nkp^{-1/2}$ reveals that $\bar{g}_{S,j}$ (without double-normalization) is proportional to $\nkp^{-1}$.

This strong scaling is mitigated by detectability, which becomes worse as $\nkp$ increases.
For the non-centrality, the test statistic $\|\vv{r}\|_2^2$ has mean $2\nkp - m$ under $H_0$ \eqref{eq:q-nofault} and $2\nkp - m + \lambda^2$ under $H_S$ \eqref{eq:q-fault}, so the two distributions are separated in mean by exactly $\lambda^2$.
Both components that determine $\bar{\lambda}$, including the detection threshold $\tau$ from \cref{eq:choose-tau} and the missed-detection target from \cref{eq:lambda-bar}, scale with the $H_0$ standard deviation $\sqrt{2(2\nkp - m)} \propto \nkp^{1/2}$.
Hence, $\bar{\lambda}^2 \propto \nkp^{1/2}$ and $\bar{\lambda} \propto \nkp^{1/4}$.
Multiplying yields
\begin{equation}
  \PL_S = \bar{g}_{S,j}\, \bar{\lambda} \propto \nkp^{-3/4}.
\end{equation}

The fault term decays faster than the nominal term, so the nominal contribution governs the asymptotic $\PL_j \propto \nkp^{-1/2}$, with the fault share contributing a small offset at modest $\nkp$ and vanishing thereafter.

\subsection{Position-only vs.\ full-pose estimation}
Finally, we consider the effect of estimating position rather than the full pose.
When the attitude of the aircraft is supplied externally, for example, from an inertial prior, the optimizer only estimates the position $\vv{p} \in \reals^3$ from the same residuals.
If the attitude is treated as known, this position-only problem is the constrained version of the full-pose problem with the orientation parameters fixed.
Removing free parameters from a least-squares problem cannot increase the covariance of the remaining parameters, so $\sigma_{\beta_j}^\mathrm{pos} \leq \sigma_{\beta_j}^\mathrm{full}$ for each position axis $j$.
The inequality is strict whenever the position and orientation columns of $\HH$ are correlated.
This effect is most pronounced near head-on viewing geometries, where along-track translation and pitch induce nearly identical pixel motion.
The residual degrees of freedom also increase from $2\nkp - 6$ to $2\nkp - 3$, which slightly increases $\tau$ and $\bar{\lambda}$; however, this logarithmic effect is small compared to the reduction in pose covariance.

\section{Experiments}\label{sec:experiments}
We demonstrate the algorithm presented on a representative runway scenario, characterizing how PLs vary with viewing geometry and keypoint count.

\subsection{Simulation Setup}\label{sec:sim_setup}

We consider a $50\,\mathrm{m}$ wide, $3\,\mathrm{km}$ long runway with keypoints at the four corners.
The simulated camera ($4000 \times 3000\,\mathrm{px}$, $f = 25\,\mathrm{mm}$, pixel size $3.4\,\mu\mathrm{m}$) is placed $1\,\mathrm{km}$ from the threshold at $150\,\mathrm{m}$ elevation, centered on the runway.
We add i.i.d.\ Gaussian noise with $\sigma = 1\,\mathrm{px}$ to each pixel coordinate, producing $2\nkp = 8$ measurements and $m = 6$ states, yielding $2\nkp - m = 2$ residual degrees of freedom.
The PnP problem is solved via Levenberg--Marquardt, with a log-transform on the along-track distance and elevation to enforce sign constraints.
Unless stated otherwise, we use a continuity-risk budget $C_\REQ = 10^{-5}$, an integrity-risk budget $I_\REQ = 10^{-5}$, and a per-keypoint failure prior $\PP_\kp = 10^{-4}$, with $\smax = 1$ (single-keypoint hypotheses) for the four-corner geometry where multi-fault cases are not geometrically detectable.
These inputs are fed into \Cref{alg:pl}: $C_\REQ$ sets the residual threshold $\tau$, the equal-split allocation $I_{H_0} = I_\REQ/2$ sets the no-fault multiplier $k_{H_0}$, and the prior-weighted allocation of the remaining $I_\REQ - I_{H_0}$ across $\dutchcal{S}_\mon$ sets the single non-centrality scale $\bar{\lambda}$.

\subsection{Geometric Sensitivity}\label{sec:geometry}
The PL expressions derived above depend on the whitened observation matrix $\HH$, which itself is determined by the camera--runway geometry.
In this section, we analyze how PLs vary with distance to runway and measurement redundancy.

\subsubsection{Distance to runway}
\begin{figure}[t]
  \centering
  \begin{subfigure}{0.524\columnwidth}
    \centering
    \includegraphics[width=\linewidth]{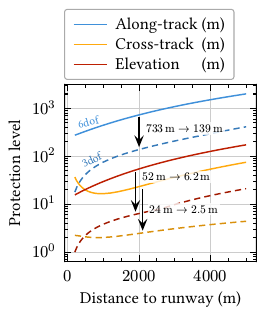}
    \caption{Position (PL in m).}
    \label{fig:pl_position}
  \end{subfigure}\hfill
  \begin{subfigure}{0.476\columnwidth}
    \centering
    \includegraphics[width=\linewidth]{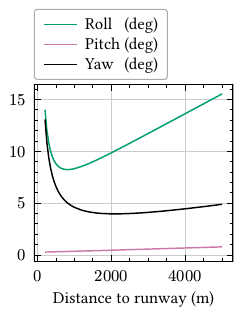}
    \caption{Orientation (PL in deg).}
    \label{fig:pl_orientation}
  \end{subfigure}
  \caption{Position and orientation PLs vs.\ distance to runway ($\nkp = 4$, $\sigma = 1\,\mathrm{px}$). In the position panel, solid curves show full-pose PLs, while dashed curves show position-only PLs.}
  \label{fig:pl_distance}
\end{figure}

As the aircraft approaches along a nominal $3^\circ$ glide slope, the keypoints spread across the image, increasing the sensitivity of the pixel measurements to pose changes.
\Cref{fig:pl_position,fig:pl_orientation} show the resulting position and orientation PLs as a function of distance for $\nkp = 4$ runway corners and $\sigma = 1\,\mathrm{px}$ noise.
The along-track PL dominates at all ranges.
For the full pose-estimation problem, the along-track PLs exceed $700\,\mathrm{m}$ at $2\,\mathrm{km}$, a consequence of the near-degenerate geometry for depth estimation from a head-on viewpoint.
Cross-track and elevation PLs are an order of magnitude tighter ($24$--$52\,\mathrm{m}$), reflecting the better lateral and vertical observability.
Orientation PLs (\Cref{fig:pl_orientation}) follow a similar trend: pitch is best constrained ($< 0.5^\circ$), while roll and yaw PLs remain above $4^\circ$ due to the limited baseline from only four coplanar points.

For the position-only PLs, the results change dramatically.
Although along-track still dominates, all PLs are reduced by approximately one order of magnitude, with the along-track PL at $2\,\mathrm{km}$ of below $140\,\mathrm{m}$, and cross-track and elevation PLs of approximately $2.5\,\mathrm{m}$ and $6\,\mathrm{m}$, respectively.
We attribute this gap to weak roll and yaw observability, and additional features constraining those axes could close it.

\subsubsection{Effect of measurement redundancy}
\begin{figure}[t]
  \centering
  \includegraphics[width=\columnwidth]{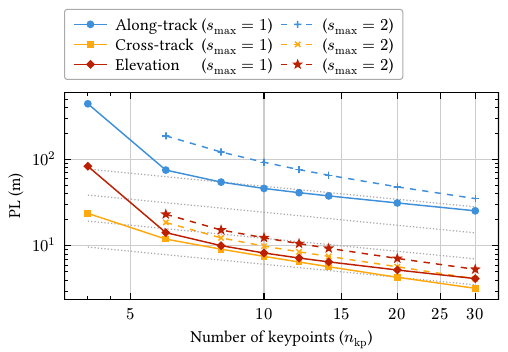}
  \caption{Position PLs vs.\ number of keypoints, log-log axes. Solid curves show $\smax = 1$ monitored hypotheses (single-keypoint faults), and colored dashed curves show $\smax = 2$ (single + double-keypoint faults). The gray dashed reference $\propto \nkp^{-1/2}$ is anchored at the along-track $\smax = 1$ PL at $\nkp = 8$.}
  \label{fig:dof_pls}
\end{figure}

With $\nkp = 4$ keypoints and $m = 6$ states, the residual has only $2\nkp - m = 2$ degrees of freedom, severely limiting the fault detection power.
Adding keypoints along the runway edges increases both the geometric information content and the residual degrees of freedom.
\Cref{fig:dof_pls} plots the position PLs against $\nkp$ on log-log axes, with the asymptotic slope $\nkp^{-1/2}$ of \Cref{sec:characteristics} overlaid as a gray reference. Past the degenerate $\nkp = 4$ point, where the residual has only $2$ degrees of freedom, the empirical PLs show the predicted asymptotic behavior. Both PL terms contribute jointly here. The fault term $\bar{g}_{S,j}\, \bar{\lambda} \propto \nkp^{-3/4}$ is initially the larger of the two and decays faster, while the nominal term $k_{H_0}\, \sigma_{\beta_j} \propto \nkp^{-1/2}$ governs the asymptote. The along-track PL drops from over $400\,\mathrm{m}$ ($\nkp = 4$) to below $26\,\mathrm{m}$ ($\nkp = 30$) for $\smax=1$. Cross-track and elevation follow the same exponent with smaller PLs due to better lateral and vertical observability.

The dashed curves in \Cref{fig:dof_pls} overlay the $\smax = 2$ PLs, where double-keypoint fault hypotheses ($|S| = 2$) are monitored. Two-fault monitoring is geometrically admissible only for $\nkp \geq 5$, because the residual must span the fault subspace, so $|S| \leq \nkp - 3$ at $m = 6$. 
Monitoring the larger fault subspace inflates the worst-case slope $\max_S \bar{g}_{S,j}$; however, the relative penalty shrinks as $\nkp$ grows, because additional measurements both tighten individual slopes and dilute the prior mass remaining in the unmonitored $|S| > 2$ tail.

\section{Conclusion}\label{sec:conclusion}
We presented a complete integrity monitoring framework for vision-based aircraft pose estimation.
By adapting residual-based RAIM to the nonlinear PnP problem, we derived analytic protection levels that bound the worst-case undetected pose error across all considered fault hypotheses, covering either the 3-DOF aircraft position or the full 6-DOF aircraft pose.

The characterization established two asymptotic facts.
First, PL scales linearly with the keypoint noise standard deviation, a relation that highlights the importance of improving keypoint prediction uncertainty.
Second, the nominal-noise term decays as $\nkp^{-1/2}$ and the fault term as $\nkp^{-3/4}$, so the nominal contribution governs the asymptotic behavior with increased number of keypoints. In the small- to moderate-$\nkp$ regime relevant to runway corner detection, both terms remain comparable and the empirical PL tracks the $\nkp^{-1/2}$ reference, with along-track PLs empirically dropping from over $400\,\mathrm{m}$ ($\nkp = 4$) to roughly $32\,\mathrm{m}$ ($\nkp = 20$) at a baseline $1\,\mathrm{km}$ distance to the runway.

Finally, we showed empirically that full pose-estimation yields PLs about an order of magnitude larger than position-only estimation, which we attribute to the poor detectability of orientation faults.
Use of these bounds as operational PLs requires bounding the deployed detector's per-keypoint fault rate and validating its uncertainty calibration.
Future work will validate the framework on real runway imagery from the LARD dataset \cite{ducoffe2023}, calibrate $\PP_\kp$ from prediction failure-mode statistics, and integrate the integrity monitor into a full approach and landing navigation architecture, including fault exclusion.

\section*{Acknowledgment}
This research was generously supported by A$^3$ by Airbus. The authors thank Stephen Dresselhaus-Marais and Sydney Katz for helpful discussions, and Todd Walter and Juan Blanch for their valuable feedback.

\bibliographystyle{IEEEtran}
\bibliography{sislstrings,refs}

\end{document}